\documentclass[]{interact}
\usepackage{soul,color}
\usepackage{epstopdf}% To incorporate .eps illustrations using PDFLaTeX, etc.
\usepackage{subfig}% Support for small, `sub' figures and tables
\usepackage{url}
\usepackage{natbib}% Citation support using natbib.sty
\bibpunct[, ]{(}{)}{,}{a}{}{,}% Citation support using natbib.sty
\renewcommand\bibfont{\fontsize{10}{12}\selectfont}% Bibliography support using natbib.sty

\theoremstyle{plain}% Theorem-like structures

\theoremstyle{definition}

\theoremstyle{remark}

\begin{document}

\articletype{Research Article}

\title{DeepSSN: a deep convolutional neural network to assess spatial scene similarity}

\author{
\name{Danhuai Guo\textsuperscript{a,b}\thanks{Corresponding Authors: Danhuai Guo (email: gdh@buct.edu.cn) and Song Gao (email: song.gao@wisc.edu)}, 
Shiyin Ge\textsuperscript{b},
Shu Zhang\textsuperscript{b},
Song Gao\textsuperscript{c},
Ran Tao\textsuperscript{d},
Yangang Wang\textsuperscript{b}}
\affil{\textsuperscript{a}Spatio-Temporal Data Intelligence Lab, College of Information Science and Technology, Beijing University of Chemical Technology, Beijing, China; 
\textsuperscript{b}Computer Network Information Center, Chinese Academy of Sciences, Beijing, China;
\textsuperscript{c}Geospatial Data Science Lab, Department of Geography, University of Wisconsin-Madison, WI, USA;
\textsuperscript{d}School of Geosciences, University of South Florida, FL, USA}
}
\maketitle

\begin{abstract} %% within 200 words
\footnote{Final Version is available on the journal of \textit{Transactions in GIS}.} Spatial-query-by-sketch is an intuitive tool to explore human spatial knowledge about geographic environments and to support communication with scene database queries. However, traditional sketch-based spatial search methods perform insufficiently due to their inability to find hidden multi-scale map features from mental sketches. In this research, we propose a deep convolutional neural network, namely Deep Spatial Scene Network (DeepSSN), to better assess the spatial scene similarity. In DeepSSN, a triplet loss function is designed as a comprehensive distance metric to support the similarity assessment. A positive and negative example mining strategy using qualitative constraint networks in spatial reasoning is designed to ensure a consistently increasing distinction of triplets during the training process. Moreover, we develop a prototype spatial scene search system using the proposed DeepSSN, in which the users input spatial query via sketch maps and the system can automatically augment the sketch training data. The proposed model is validated using multi-source conflated map data including 131,300 labeled scene samples after data augmentation. The empirical results demonstrate that the DeepSSN outperforms baseline methods including k-nearest-neighbors, multilayer perceptron, AlexNet, DenseNet, and ResNet using mean reciprocal rank and precision metrics. This research advances geographic information retrieval studies by introducing a novel deep learning method tailored to spatial scene queries. 
\end{abstract}


\section{Introduction}
% Spatial scene similarity assessment & Spatial-query-by-sketch
Spatial-query-by-sketch (SQBS), i.e. searching for spatial scenes similar to a sketch map drawn by users, is a promising way to query a geospatial database \citep{egenhofer1997query} and plays an important role in geographic information retrieval (GIR) \citep{larson1996geographic,jones2008geographical,janowicz2011semantics}. Due to the intuitiveness of sketch maps and the ubiquity of drawn-friendly technology and devices, such as laptops and smartphones, SQBS was recognized as a promising approach to interacting with Geographic Information Systems (GIS). Several prototypes such as the SQBS system\footnote{\url{http://sqbs.org/}} and SketchMapia\footnote{\url{http://giv-sketchmapia.uni-muenster.de:8080/sketchMapiaInterface/index.php}} \citep{schwering2010sketchmapia} have been developed in scientific research. The SQBS systems have potential applications in (a) urban design, where it helps planners find cities with similar layout patterns compared to a draft sketch design; (b) location query, where the user’s inputs include a complicated spatial configuration beyond keywords and spatial proximity as current map search systems provide. For example, a traveler may want to find a hotel near a lake and a subway station, but far away from a noisy market; (c) criminal investigation, for example, an investigator needs to find or rebuild the scene of a crime according to oral confessions or testimony, and (d) housing service platforms, where sketch-based models can help agents provide diversified services such as searching for specified styles of spatial scenes or identifying appropriate houses according to personalized requirements, i.e. surrounding facilities, which can be expressed through a sketch. However, there are still many challenges needed to be addressed before general SQBS systems can be realized into those applications including semantic ambiguity, uncertainty of sketches, and implicit spatial information.

% traditional methods
The performance of SQBS relies on the spatial scene similarity assessment between a sketch query and the candidate spatial scenes \citep{wang2011empirical}. Most existing methods of spatial scene similarity assessment consider geometric similarity, semantic similarity, and the similarity of spatial relations among geographic objects. A common way to build a similarity-based spatial scene searching system is to first extract the spatial configuration from a sketch or an icon-based map, and measure the similarity between the retrieved spatial scene from a sketch and candidate scenes from spatial database based on their spatial attributes such as type, size and shape as well as their spatial relationships including topology, direction and proximity \citep{wang2015invariant}. However, the performance of those systems is sensitive to the accuracy of spatial object extraction and spatial relation retrieval. Moreover, some implicit but important features and relationships of the spatial scenes are often ignored in those methods, since they only focus on the similarity between the paired spatial objects. The implicit features and patterns are mostly hidden among the spatial configuration of multiple objects and their complex relations. As the number of spatial objects increases, the computation time increases exponentially.

% Deep learning similar & difficulty
To better capture the complex features and implicit spatial patterns of digital objects, deep learning models such as convolutional neural networks (CNNs) and recurrent neural networks (RNNs) have recently been developed, leading to tremendous advancement in image classification \citep{krizhevsky2012imagenet}, object detection \citep{ren2015faster}, scene recognition \citep{donahue2014decaf}, and many other computer vision tasks \citep{lecun2010convolutional}. New studies have emerged in geospatial artificial intelligence (GeoAI) \citep{hu2019geoai,hu2019artificial,janowicz2020geoai}, such as automatic terrain feature identification \citep{li2018automated}, species mapping \citep{wan2019small}, contemporary vector data alignment with historical maps \citep{duan2017automatic}, spatial interpolation analysis \citep{zhu2019spatial}, understanding place semantics and human-environment interactions \citep{zhang2018representing,kang2019extracting}, parking violation prediction \citep{gao2019predicting}, and living environment classification and building function interpretation \citep{srivastava2018multilabel,yan2018xnet, liu2019inside}. In these various tasks, deep learning models essentially achieve an automatic feature engineering and encoding of the underlying features, including local and global characteristics. However, the spatial scene similarity assessment differs from most regular vision tasks in several ways. On the one hand, spatial scenes contain multi-scale spatially explicit contextual information, which is critical in GeoAI tasks \citep{janowicz2020geoai,mai2020multi,yan2018xnet}. On the other hand, image classification requires huge number of labeled images as training samples, whereas the spatial scene similarity assessment often faces the issue of an insufficient number of labeled sketch maps. Furthermore, a sketch map usually contains censored and noisy information, which increases difficulty to the extraction of generic features.

% propose method
To overcome these problems, we propose a novel Deep Spatial Scene Network (DeepSSN) to extract features to represent multi-scale spatial scenes and assess the spatial scene similarity. DeepSSN is based on a specified multi-layer convolutional neural network. To accommodate the multi-scale geographic properties, a spatial pyramid pooling (SPP) layer \citep{he2015spatial} (see Section \ref{sec:embedding-network} for more details) is adopted to generate a fixed-length representation, which retains the multi-scale geometric information with an arbitrary image size or scale. 
In addition, a deep convolutional network is trained to optimize the feature representation with a triplet loss function (see Section \ref{sec:triplet-loss} for more details) that serves as a distance metric for searching similar spatial scenes. During the model training, the selection of appropriate triplets (i.e., pairs of anchor, positive, and negative image samples) is crucial to the performance of DeepSSN. Therefore, we come up with an effective strategy to mine positive and negative samples, which ensures a consistent distinction of triplets during the training process. To collect a sufficient quantity of sketch map data as training samples, we build a sketch-based spatial search system to collect the users' operation log to build a labelled sample dataset. Users' satisfaction to the matched results is further ranked and adopted as the ground truth. The experiment results demonstrate that the DeepSSN can effectively assess spatial scene similarity and that it outperforms several existing baseline methods in both precision and convergence time.

% contribution
The major contributions of this paper include: 
\begin{enumerate}
\item A novel DeepSSN model, which can automatically learn a unified representation of spatial scenes and improving the extraction of implicit features and relationships for similarity search and retrieval tasks, is proposed.
\item A computationally-efficient triplet mining strategy for spatial scenes, which can significantly improve the GIR performance of DeepSSN, is proposed.
\item Based on the DeepSSN model, a sketch-based spatial query system is developed to support practical applications related to spatial scene similarity assessment.
\end{enumerate}

% structure information
The remainder of this paper is organized as follows. In section 2, we first review related works on sketch maps and spatial scene search in GIScience and computer sciences. In section 3, we introduce the proposed DeepSSN model in detail. In section 4, we present the datasets and an augmentation strategy used in the study. In section 5, we test our method and compare the model performance with other spatial scene search baseline methods. In section 6, we discuss the impacts of embedding vector dimensions of spatial scenes and training data sparsity issue, as well as the limitations in our method. Finally, we conclude this work with major findings followed by our vision for future work and insights into the advancement of deep learning in GeoAI in section 7.

\section{Related work}
Sketch maps are often used as an investigation technique that involves human spatial cognition. A number of methods have been proposed in previous literature to represent and recognize spatial scenes in both qualitative and quantitative ways \citep{billinghurst1995use, tu2007digital, chipofya2011towards, forbus2011cogsketch, schwering2014sketchmapia}. For instance, \citet{enenhofer1996spatial} proposed an intuitive interaction method for spatial search in GIS that allows a user to formulate a spatial query by a sketch map. Several approaches \citep{egenhofer1997query, haarslev1997querying, winter2000location, caduff2005geo, liang2005sketch} have successfully captured spatial attributes and spatial relations between depicted sketches. 

Meanwhile, researchers have taken great strides in developing theoretical frameworks and practical SQBS systems. For example,  \citet{haarslev1997querying} designed a visual query system VISCO, which provides a sketch-based query language for approximate spatial queries with geometrical and topological constraints. \citet{nedas2008spatial} proposed a similarity measure for comparing two spatial scenes by identifying cognitive similarities and the relations among spatial objects. \citet{wallgrun2010qualitative} developed a matching approach named as Qualitative Matching. It represents a sketch map as a set of qualitative constraint networks (QCN) and uses qualitative spatial reasoning to prune the search space. \citet{inproceedings} implemented the QCN in their toolbox \textit{SparQ}. In addition, \citet{schwering2014sketchmapia} proposed a qualitative representation system \textit{SketchMapia} for aligning information from sketch maps and the metric basemap in a volunteered geographic information (VGI) platform \citep{goodchild2007citizens}. The system takes images of sketch maps drawn with pencil rather than using digital ink. In addition, with the increasing availability of VGI data, a viewpoint-based semantic similarity GIR system was developed using OpenStreetMap \citep{ballatore2012holistic}. Recently, \citet{wang2015invariant} presented a cognitive approach to identifying spatial objects (including landmarks, street segments, junctions, and city blocks) and invariant spatial configurations in sketch maps.

Being limited by capacities of spatial relationship reasoning methods, these classic sketch-based spatial query methods show inefficiencies in retrieving implicit spatial query information. Sketch-based spatial scene search is similar to the sketch-based image retrieval, where significant progress has been made recently due to the introduction of deep distance metric learning \citep{weinberger2009distance}. The core of deep leaning based image match is to find a loss function to evaluate the fit degree of a learning model to the given data. Triplet loss, which is a loss function of distance metric learning \citep{weinberger2009distance}, has already been widely used in several image retrieval and image-based person re-identification (ReID) tasks, in which the semantic, visual, and spatial similarity information is used \citep{wang2014learning, schroff2015facenet}. It helps to minimize the \begin{math}L2\end{math}-distance (the Euclidean norm) between images of the same category and restricts a distance margin between images of different categories. In particular, the recent successful image retrieval approaches employ the triplet loss for metric learning and designate a deep convolution neural network as a feature extractor to encode images with a comprehensive embedding layer \citep{liu2016deep, paisitkriangkrai2015learning}. The embeddings (i.e., multi-dimensional
vectors) help answer the question of ``How similar are the two images", which could also be used in many other tasks such as objects clustering and verification \citep{hermans2017defense}. A triplet mining strategy tailored to spatial scenes is developed in this research. 

In addition, \citet{VincentSpruyt2018} learned from the Word2Vec model and developed a deep learning based method Loc2Vec to encode spatial relation and semantic similarity. When the user inputs a location, the Loc2Vec would output an embedding that captures high-level semantics of the input location. Based on this idea, \citet{Samano2020} presented a novel approach to geolocalising panoramic images on a 2D cartographic map by learning a low dimensional embedding space, which allows a comparison between an image captured at a location and local neighbourhoods of the map. In spite of similar methods were employed such as CNN and a triplet loss function by Loc2Vec and DeepSSN, our proposed DeepSSN shows significant differences to Loc2Vec. DeepSSN focuses on assessing the similarity between two spatial scenes, while Loc2Vec focuses on the construction and visualization of embeddings, which is only the representational learning step of the model for similar image search. In addition, our proposed work supports sketch-based spatial scene similarity query.

\section{Methods}
% Overview
\begin{figure}[h]
  \centering
  \includegraphics[width=\linewidth]{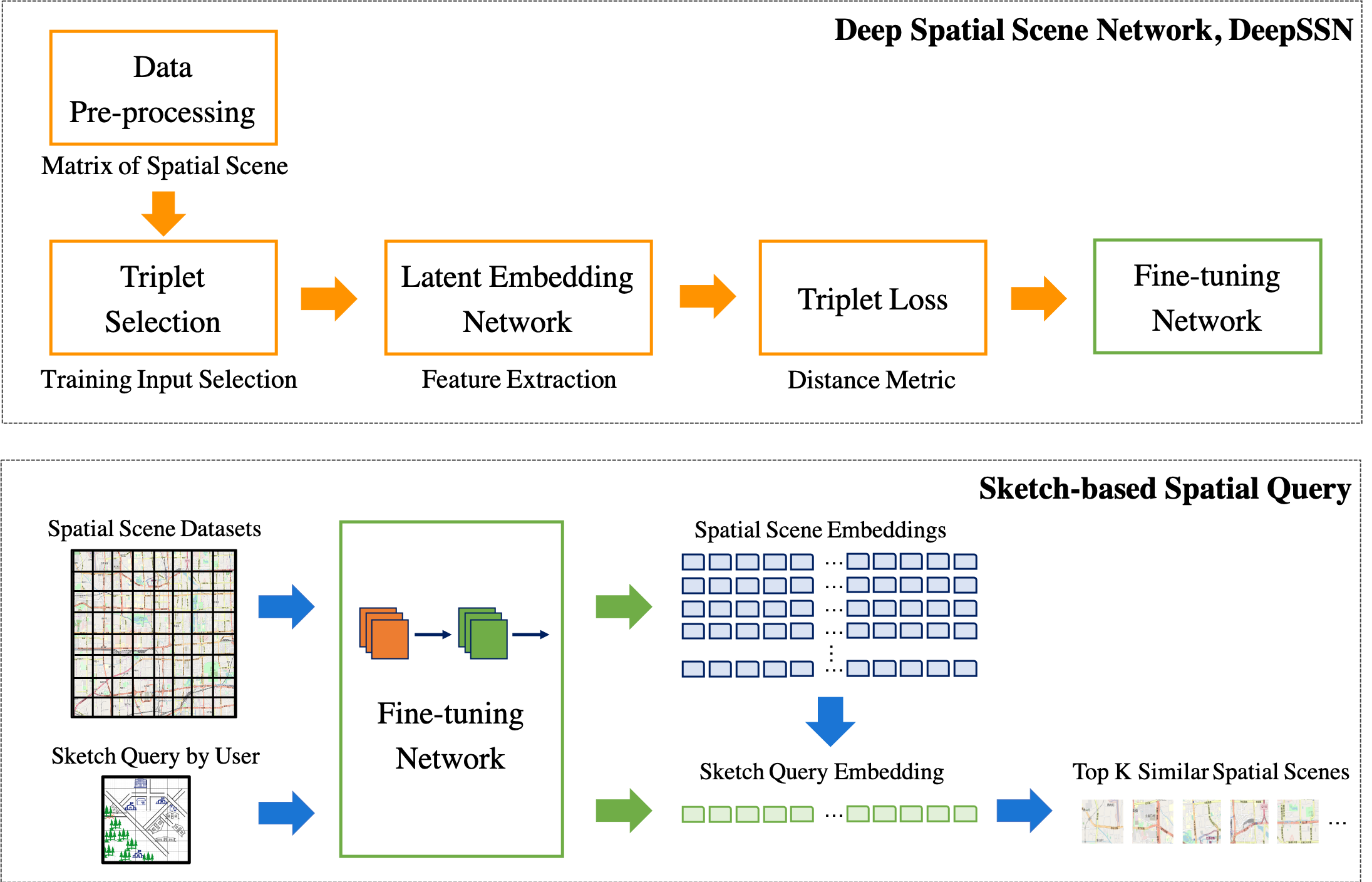}
  \caption{The overall framework of DeepSSN.}
  \label{fig:overall}
\end{figure}

In the following section, we present a deep convolutional neural network based spatial scene similarity search system called DeepSSN. The overall framework of DeepSSN is shown in Figure \ref{fig:overall}. DeepSSN consists of two parts: a CNN for spatial scene similarity assessment and a sketch-based spatial query component. A set of robust and distinctive features of a spatial scene are crucial for the task of spatial scene search. The search can usually be achieved by comparing the geometric attributes and spatial relationships of a sketch map and the spatial scenes in a database. Thus, the process of spatial scene search is similar to a retrieval task, where deep learning and triplet loss have been introduced with significant success. The CNN for spatial scene similarity assessment is designed to learn an embedding mapping function from a spatial scene to a compact Euclidean embedding space where distance metrics serve as a measure of spatial scene similarity. Meanwhile, the embedding mapping keeps the representative features of spatial scenes. Once the compact embedding of spatial scenes has been learned, the similar scene search associated with each sketch query becomes a rank-based k-nearest neighbor (k-NN, or KNN) classification problem \citep{roussopoulos1995nearest,weinberger2006distance}. The following section describes the details of DeepSSN and how it can learn efficiently in the sketch-based spatial query task.
Furthermore, we provided the open-source code for this research on Github at the link \url{https://github.com/SayaZhang/Spatial-Query-by-Sketch} to support replicability and reproducibility in geospatial research \citep{kedron2019rr,janowicz2020geoai}.

\subsection{Latent Embedding Network}\label{sec:embedding-network}
\begin{figure}
	\centering
	\includegraphics[width=\linewidth]{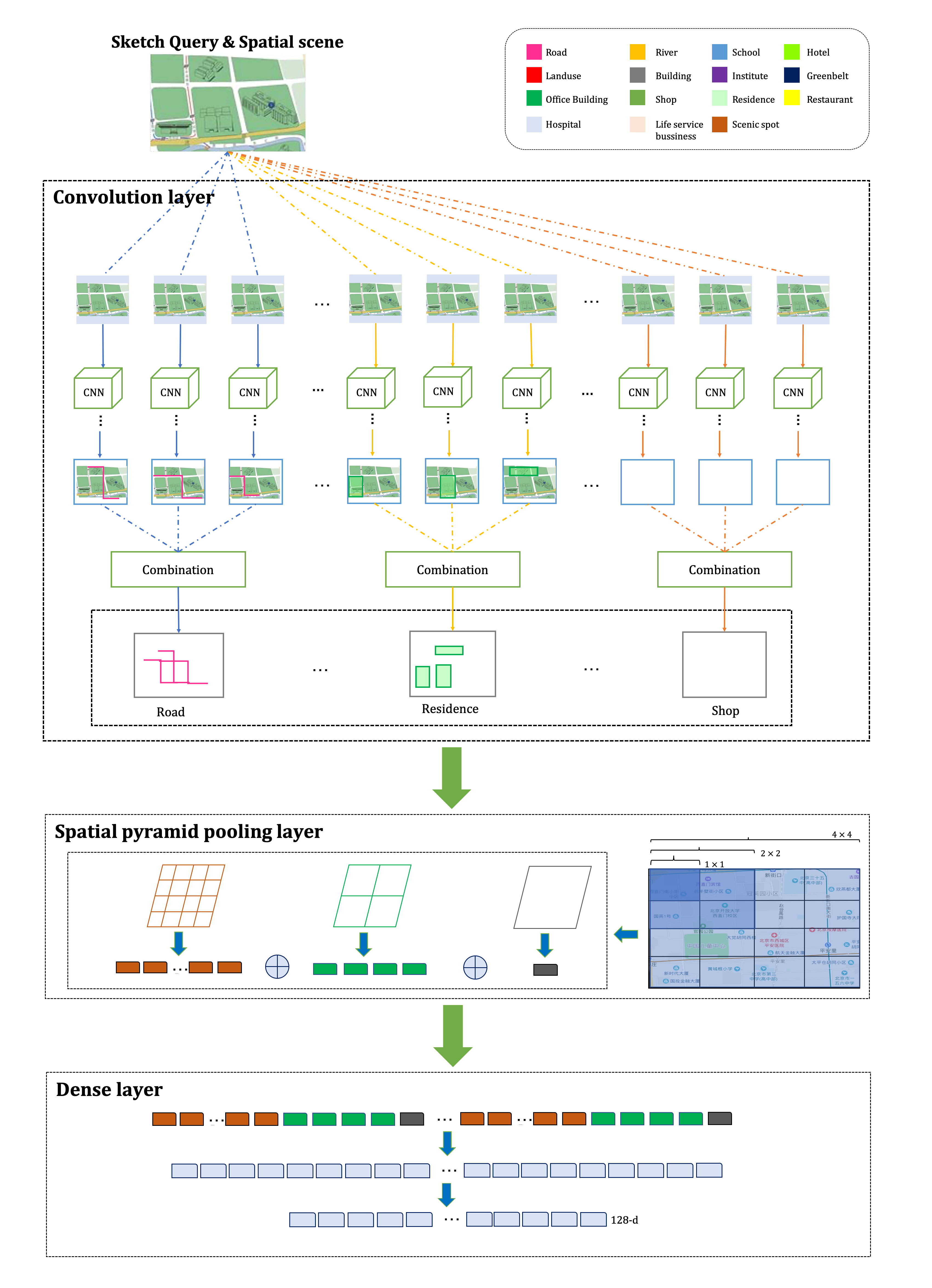}
	\caption{The latent embedding network architecture in DeepSSN, which is composed of a sequence of convolution layers, a spatial pyramid layer and dense layers. The feature extraction architecture is used to transform the input high-dimensional spatial scenes (and sketch queries) consisting of various types of geographic objects to the Euclidean space of representative embedding. }
	\label{fig:Embedding}
\end{figure}

As discussed earlier, the similarity measurement weighs the types of geographic objects and their spatial relations, including topology, direction, and proximity. However, it is a great challenge to extract the hand-drawn features in sketches. A convolution operator addresses this challenge by preserving the spatial relations and localities of the input data in a latent high-level feature representation space. Given that the number of free parameters used for describing the shared weights is independent from the input's dimensionality \citep{masci2011stacked}, it is possible to design a deep convolutional neural network architecture to deal with high-dimensional real-world spatial data. Taking into account the peculiarities of feature extraction for spatial scenes, we introduce a latent embedding network based on the idea of convolution, spatial pyramid pooling and dense neural networks. The objective of the latent embedding network is to learn an embedding function \begin{math}f(.)\end{math}, which transforms the spatial scenes consisting of various types of geographic objects to a multi-dimensional Euclidean space of representative embedding. As shown in Fig. \ref{fig:Embedding}, the latent embedding neural network for transforming those spatial scenes is composed of a sequence of convolution layers, a spatial pyramid layer and fully-connected dense layers, which are described in detail in the following subsections, respectively.

\subsubsection{Convolution layers} 
The most important layers in CNNs are convolution layers and pooling layers. The former extracts local regional features using multiple filters and combines the responses by referring to the locations of local regions. The latter summarizes the feature responses. The stride for filter shift and the kernel size for processing the pooling layers are fixed. Here we encode sparse spatial correlations based on convolutional layers in spatial scene representation for a better similarity search. We employ the convolution layers as feature filters, which are adaptive to small regions and good for exploiting multi-scale localized feature interactions and constructing high-level semantic representations.

A CNN model with outstanding performance in image classification may not work well in spatial scene similarity searching with sketch queries. Due to the sparsity of training samples and complex geographic relations in spatial scenes, a large number of pixels are necessary to recognize multi-scale geographic objects and their relationships \citep{cao2015practical,ning2020choosing}. We set the size of the convolutional filter much larger than the norm (3 $\times$ 3) in image classification, specifically (11,7,5) in three convolution layers. Meanwhile, we use the convolution layer with a larger number of filter maps. This helps extract latent features in geographic space and avoid loss of information. The input matrix dimensionality of the first CNN layer is 40$\times$40$\times$15 (height, width, channels).

Additionally, we optimized the hyper-parameter setting using an efficient random search strategy rather than exhaustive grid search over a large-sized hyper-parameter configuration space \citep{Bergstra:2012:RSH:2503308.2188395}. The detailed selection of parameters can be found in Table \ref{tab:net-architectures} and in the open-source code repository mentioned above.

\subsubsection{Spatial pyramid pooling}
To extract multi-scale features, DeepSSN employs a combination of convolution and spatial pyramid pooling techniques \citep{he2015spatial}. Thus, it can capture spatial contextual information at multiple scales for a more accurate ranking result. The number of local features for a sample in a traditional bag-of-features model varies \citep{liu2012has}. To obtain a fixed-length feature representation, we need to encode and transform the local features into a encoded vector, followed by the pooling step to achieve the global-scale representation. In this work, the spatial pyramid pooling incorporates multi-scale local features by conducting the pooling operator at different spatial scales. 

The responses of all filters in each spatial scale are pooled with the max pooling throughout this research. The outputs are \begin{math}k \times M\end{math}-dimensional vectors, where $k$ and $M$ represent the number of filters and the size of bins in the last convolutional layer, respectively. We choose a three-level spatial pyramid max pooling with $4\times4,2\times2,1\times1$ bin sizes to better capture meaningful features across locations and scales in spatial scenes. Spatial pyramid pooling makes multi-scale features adaptive to the sketch query through a learning process.

\subsubsection{Fully-connected dense layers} 
The way to learn non-linear combinations of features is to use fully-connected dense layers \citep{sainath2015convolutional}. When training the DeepSSN model, we expect that, after processing a sufficient number of training cases, the fully-connected layer could learn a location-invariant function to classify and down-sample the features of the convolutional layers and the spatial pyramid pooling layer. Specifically, the first fully-connected layer receiving the output of the spatial pyramid pooling layer contains 4096 neurons, which is followed by a ReLU (i.e., rectified linear units) activation function and a dropout layer \citep{goodfellow2016deep}. The second layer contains 128 neurons, and it maps to the final embedding of the spatial scene. The final layer is fed to a ranking layer that outputs a multi-dimensional vector for each input spatial scene. We also further discuss the impact of the output embedding dimension in terms of accuracy and computation time in section 6.1.

\begin{table}[!htbp]
\centering
\caption{The network configuration of DeepSSN with hyperparameter setting.}
\label{tab:net-architectures}
\begin{tabular}{ccr}
\toprule
Layers& Operator Shape& Parameters\\
\midrule
Conv1 & feature maps=96, kernel size=11& 174,336\\
Conv2-batchNorm & feature maps=256, kernel size=7& 1,205,504\\
Conv3-batchNorm & feature maps=384, kernel size=5& 2,459,520\\
Spatial Pyramid Pooling & $4\times4,2\times2,1\times1$ max pooling& \\
Fully-connected1 & 4096 neurons, drop rate=0.2 & 33,034,240\\
Fully-connected2 & 128 neurons & 524,416\\
\bottomrule
\end{tabular}
\end{table}

% model training details
The network configuration with hyperparameter settings of the DeepSSN model is shown in Table \ref{tab:net-architectures}. We choose the triplet loss function to evaluate the information loss and update the model parameters of DeepSSN. To calculate the triplet loss, the inputs are spatial scene triplets, which are generated by the hard triplet selection strategy. One triplet contains an anchor spatial scene \begin{math}x^a_i\end{math}, a positive sketch \begin{math}x^+_i\end{math}, and a negative spatial scene \begin{math}x^-_i\end{math}, which are fed into a latent embedding network respectively with a shared network architecture and hyperparameter setting, whose output is a dense embedding of a spatial scene \citep{wang2014learning}. The details of triplet loss and triplet selection strategy are described in sections \ref{sec:triplet-loss} and \ref{sec:triplet-selection}. Based on the triplet loss, the DeepSSN evaluates the loss of a triplet and computes the distances between positive and negative spatial scenes. Using the back propagation gradients, DeepSSN learns and evaluates the violation of the ranking order of similar spatial scene embeddings so that these preceding layers can update the parameters with the aim of minimizing the triplet loss \citep{wang2014learning}. Additionally, we train the DeepSSN by using stochastic gradient descent (SGD) with a batch size of 512, a learning rate of 0.01, and 20 epochs, configured by the hyper-parameter tuning process. 

% embedding -> Top K search
 After the latent embedding network has been trained, both the sketch query and a set of spatial scene candidates are mapped into 128 dimensional embedding vectors, where distances directly correspond to a measure of spatial similarity. Then the similar spatial scene search problem becomes a k-nearest-neighbor ranking problem \citep{schroff2015facenet}, which is shown in the sketch-based spatial query part of the general workflow (in Fig. \ref{fig:overall}). The search task involves the computation of the distance between the candidate spatial scene embedding and the sketch embedding, and the ranking of spatial scene candidates by their distances using multi-dimensional vectors.

\subsection{Triplet Loss function}\label{sec:triplet-loss}
\begin{figure}[h]
  \centering
  \includegraphics[width=\linewidth]{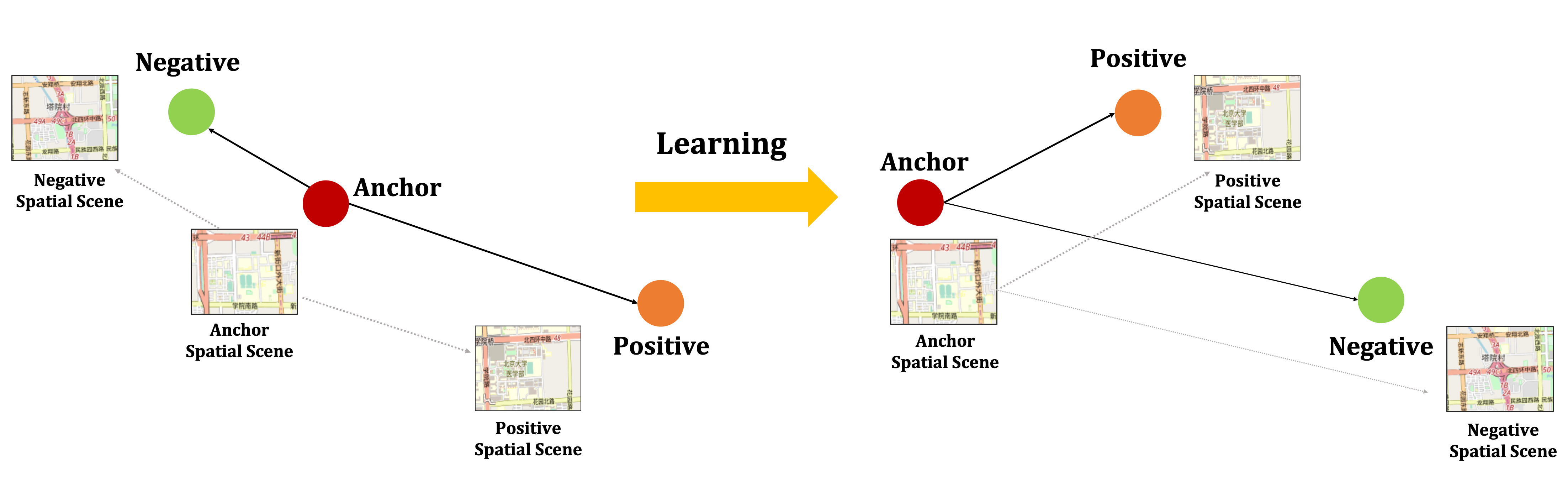}
  \caption{The triplet loss in DeepSSN minimizes the distance between an \textit{anchor} spatial scene and a \textit{positive} spatial scene that belong to the same category, while maximizing the distance of the \textit{anchor} spatial scene and a \textit{negative} spatial scene of different categories.} 
  \label{fig:tripletloss}
\end{figure}

In DeepSSN, the triplet loss plays a key role for the distance metric learning task. The triplet loss is motivated by the k-NN classification and calculated on the triplet of training samples \begin{math}(x^a_i,x^+_i,x^-_i)\end{math}, where \begin{math}(x^a_i,x^+_i)\end{math} have the same categorical labels, and \begin{math}(x^a_i,x^-_i)\end{math} have different categorical labels \citep{weinberger2009distance,schroff2015facenet}. As shown in Fig. \ref{fig:tripletloss}, a circle represents a sample and a edge represents the distance between samples. The \begin{math}x^a_i\end{math} serves as an anchor spatial scene of a triplet, where pairs of \begin{math}x^+_i,x^-_i\end{math} are the positive spatial scene samples and negative spatial scene samples, respectively. The triplet loss function assigns the distances between the anchor spatial scene and all the positive samples (i.e.,  \begin{math}x^a_i\end{math} and \begin{math}x^+_i\end{math}) closer to any negative samples (i.e.,  \begin{math}x^a_i\end{math} and \begin{math}x^-_i\end{math}) by at least a margin of \begin{math}m\end{math}, which can be expressed as:
\begin{equation}D(f(x^a_i), f(x^+_i)) + m < D(f(x^a_i), f(x^-_i))\label{eq:1}\end{equation}
\begin{equation}D(x,y) = \parallel x - y \parallel^2_2 \end{equation}
\begin{equation}\forall(f(x^a_i), f(x^+_i), f(x^-_i)) \in \Gamma\end{equation} 
where the embedding is generated by $f(x) \in R^d$, $f(.)$ is the spatial scene embedding function which maps a spatial scene into a $d$-dimensional Euclidean space. In addition, to avoid the loss function exceeding 0, the embedding is constrained in a d-dimensional hypersphere, i.e. $\parallel f(x) \parallel^2_2 = 1$. $D(.){}$ represents the squared Euclidean distance function ( $\parallel x-y\parallel^2_2$) between two vectors $x$ and $y$. $\Gamma$ has cardinality $N$ and represents the set of all possible triplets in the training data set \citep{schroff2015facenet}. 

The triplet loss is computed and minimized as follows:
\begin{equation} L = \sum^{N}_{i=1}[D(f(x^a_i), f(x^+_i)) - D(f(x^a_i), f(x^-_i)) + m]
\label{eq:4}\end{equation}
According to the definition of triplet loss, the similar spatial scene ranking problem converts to the nearest neighbor search problem in a Euclidean space; it derives a similarity ranking for the spatial scenes \begin{math}x^a_i,x^+_i,x^-_i\end{math}. 

One of the approaches to avoiding overfitting is to utilize as many sample data as possible. However, the number of possible triplets grows cubically with the number of spatial scenes \citep{zhang2019learning}. Therefore, the most important component of the triplet loss function is to select appropriate triplet samples. Since many triplets would not contribute to convergence of the model, they can be satisfied to the Eq.\eqref{eq:1} and differentiated. The following section explains our mining strategy for the triplet sample selection.

\subsection{Triplet Selection}\label{sec:triplet-selection}
It is important to select the triplets that contribute the most to the distance metric learning. Intuitively, the sketch maps of various spatial scenes with different shapes of geographic objects and at different scales provide different cues. Specifically, a spatial scene sample that has similar geographic objects but at different spatial scene categories should be defined as a hard-negative sample. Hence, the mining of hard positives and hard negatives plays an essential role in accelerating the model convergence rate.

The objective of triplet loss in distance metric learning is to minimize the distances of positive scene samples with the same label while maintaining the distances of negative scene samples with different labels above a minimum margin \citep{em2017incorporating}. Thus, ideal triplets can capture both intra-class and inter-class distances efficiently \citep{liu2016deep}. Those ``hard examples'' are often misclassified. Here, hard positives and hard negatives are defined as follows:

\begin{itemize}
\item A \textit{hard positive} sample refers to the spatial scene whose intra-class distance to an anchor spatial scene is maximum.
\item A \textit{hard negative} sample refers to the spatial scene whose inter-class distance to an anchor spatial scene is minimum.
\end{itemize}

\begin{figure}[h]
  \centering
  \includegraphics[width=\linewidth]{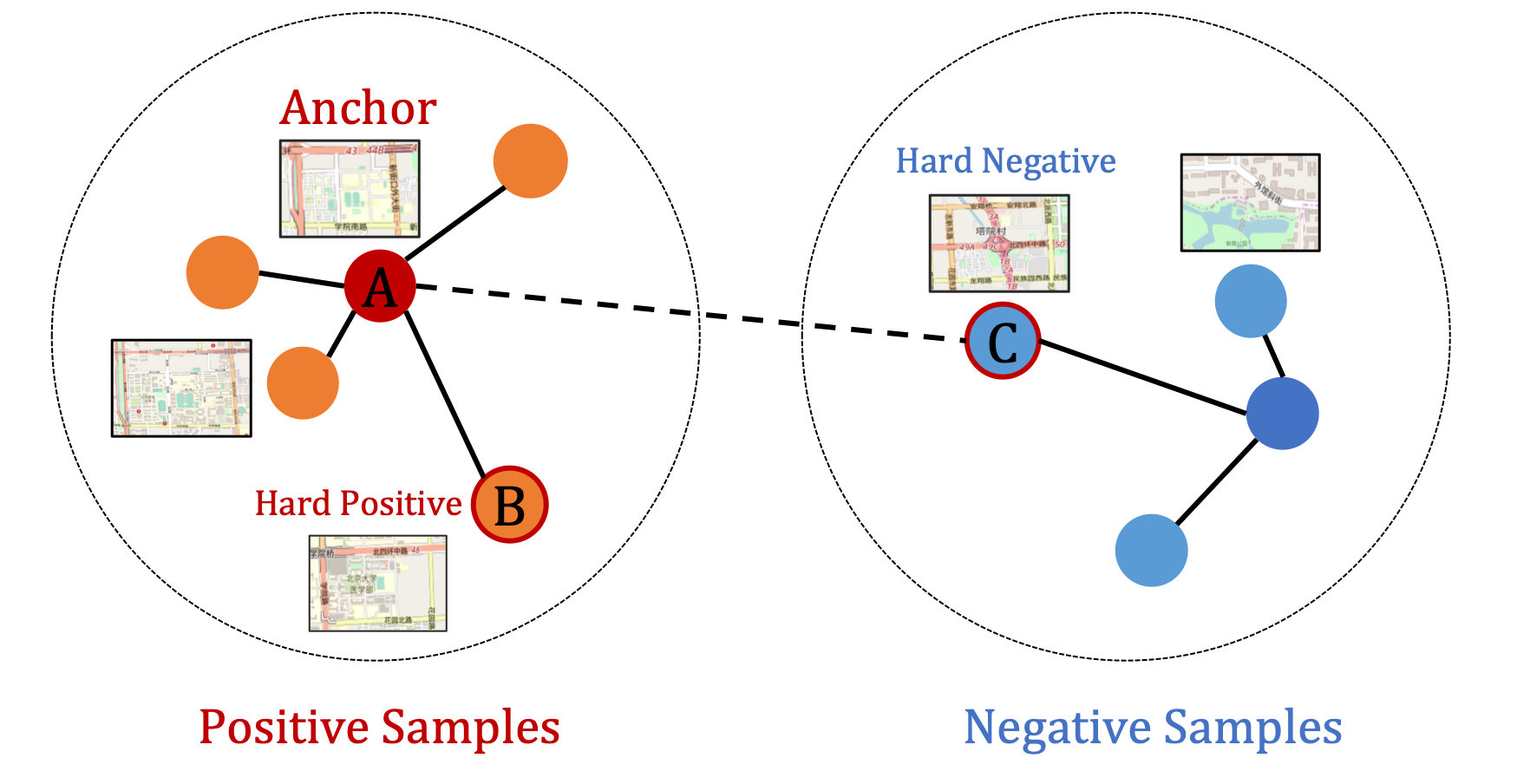}
  \caption{Generation of hard positive samples and hard negative samples of DeepSSN. The solid line refers to intra-class distance, and a dashed line refers to inter-class distance.}
  \label{fig:distance}
\end{figure}

To generate hard positive samples for an anchor spatial scene $A$, we investigate anchors that are in the same class with $A$ and find the most dissimilar spatial scene $B$. Then the pair $(A, B)$ is selected as a hard positive sample. For the hard negative samples, the anchor spatial scene $C$, which belongs to another class and is closer to $A$ than other anchors, is selected. Then the pair $(A, C)$ is employed as a hard negative sample, as illustrated in Fig. \ref{fig:distance}.

With the increase of the number of spatial scene identities, the training process would lead to local sub-optimum because it is hard to select effective hard triplets \citep{wang2017train}. Furthermore, the hardest triplets are more likely to be outliers, which often make the embedding function \begin{math}f(.)\end{math} unable to learn common patterns or associations between spatial scenes.

\begin{figure}[h]
  \centering
  \includegraphics[width=\linewidth]{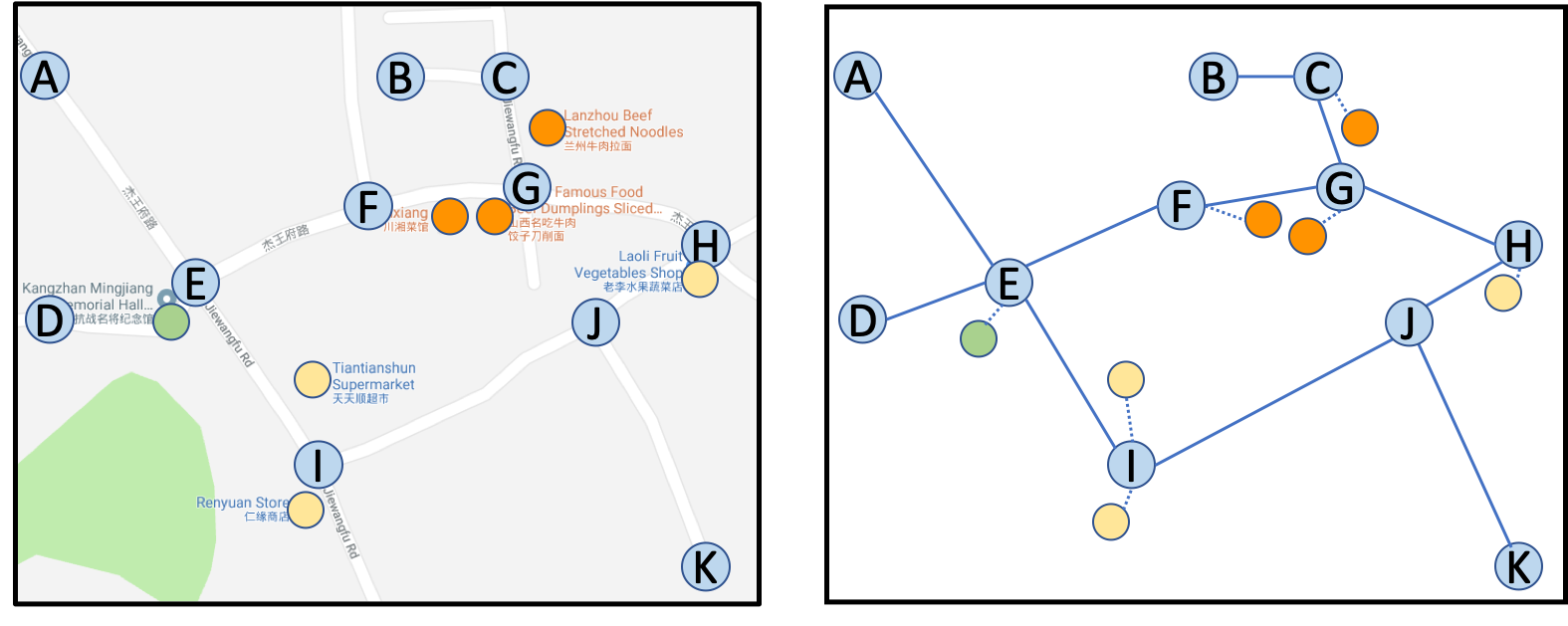}
  \caption{An example of the QCN samples. Green, yellow, and orange dots represent different types of POIs, i.e., scenic spots, supermarkets, and restaurants, respectively. The blue nodes with letters represent different road junctions while solid lines represent the road segments linking those junctions. The dash lines show the spatial closeness relation between POIs and road junctions. The left panel is an anchor spatial scene while the right panel is the corresponding qualitative constraint network interpretations.}
  \label{fig:hardPositive}
\end{figure}

To this end, we propose a novel hard triplet mining strategy for the similarity search of spatial scenes. On the one hand, for hard positives, we utilize the toolbox SparQ\footnote{\url{https://www.uni-bamberg.de/en/sme/research/sparq/}}, which applies the qualitative constraint networks (QCN) \citep{inproceedings} for each spatial scene to determine positive samples \begin{math}x^+_i\end{math} with regard to the topological, directional, and proximity relationships among objects in the scenes \citep{dylla2006sparq}. The nodes and the edges represent the objects in the scene and their relational constraints respectively in QCN. An example spatial scene for extracting the QCN from geographic objects is illustrated in Fig. \ref{fig:hardPositive}. After the qualitative interpretations are extracted, we match similar spatial scenes with similar qualitative interpretations in a spatial database as a coarse-grained positive result. Then, we rank and select positive spatial scenes from the matching result of QCN in a fine-grained way, with the selected scenes used as hard positive samples. Meanwhile, a graphic user interface was provided to support user's data labelling, which is a promising data source for selecting positive samples (more details in section \ref{sec:data}).

On the other hand, the hard negatives are expected to be similar to the anchor spatial scene but belong to different class labels. This is due to the fact that a spatial scene contains lots of geographic objects with typical semantic information, i.e., point of interest (POI), types of functional regions \citep{gao2017extracting,yan2017itdl}. The multiple categories of POIs of each spatial scene are built as a multi-dimensional vector via a global one-hot encoder. The k-NN algorithm is adopted to find the k-nearest spatial scenes in the POI feature space, thus the k-similar spatial scenes that are most similar to the anchor scene but with different labels are regarded as negative samples \begin{math}x^-_i\end{math} in the triplet. Similar to the FaceNet \citep{schroff2015facenet}, we consider several nearest \begin{math}x^-_i\end{math} samples instead of only the nearest one, to minimize the chance of including errors due to potential outliers.

\section{Training Dataset and Processing} \label{sec:data}
\subsection{Data Acquisition and Integration}

We implement an experiment to evaluate the proposed DeepSSN model using OpenStreetMap (OSM) data \citep{haklay2008openstreetmap} and large-scale commercial map data from Esri for the same area in Beijing, China. Both data sources contain various geospatial features (as shown in Table \ref{tab:geographicfeatures}) including roads, stations, buildings, land uses, and so on. Previous research has shown that OSM data from volunteers and community organizations can be used with confidence for various cartographic and mapping purposes \citep{kounadi2009assessing} in developed countries, but that the quality might be poor in developing counties \citep{haklay2010good,neis2012street,barron2014comprehensive,arsanjani2015quality,tian2019analysis}. Therefore, these two sources can complement each other after data conflation \citep{goodchild2018reimagining}.

%More specifically, the quality elements including length completeness, name accuracy, and positional accuracy are found desirable in several countries \citep{haklay2010good,neis2012street,barron2014comprehensive,arsanjani2015quality,tian2019analysis}.

\begin{figure}[h]
  \centering
  \includegraphics[width=\linewidth]{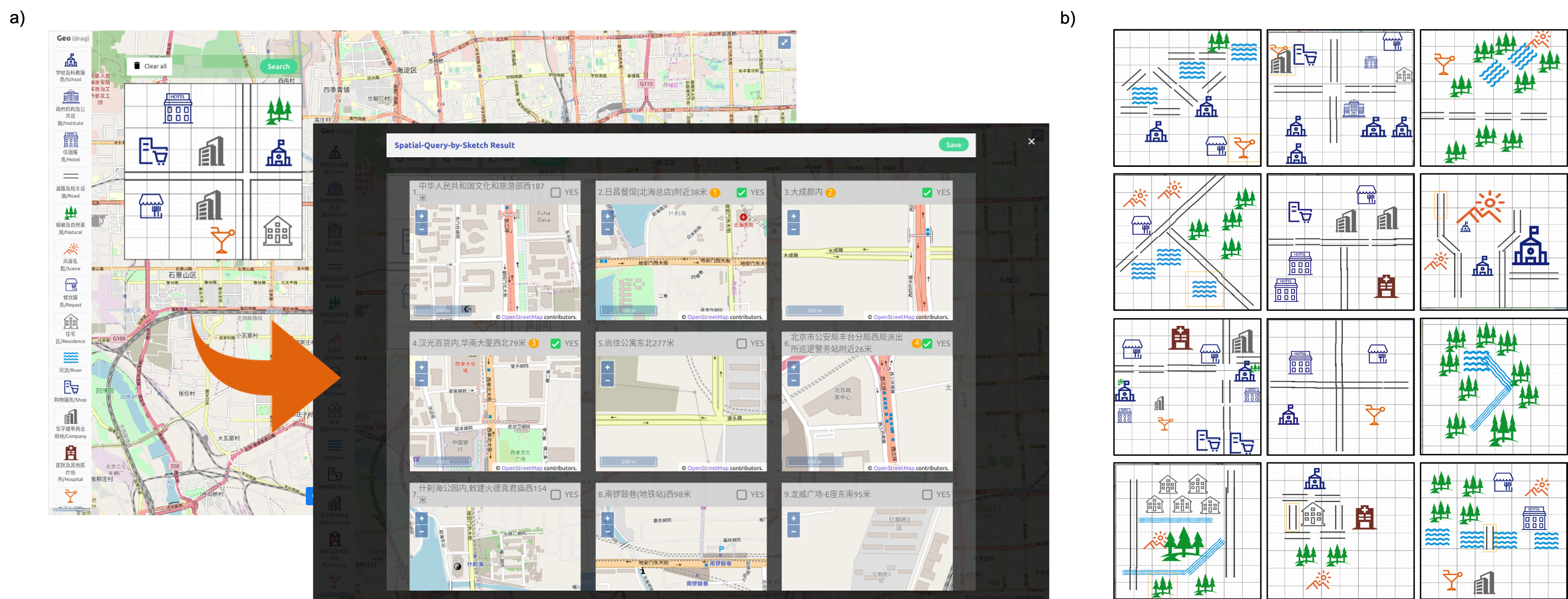}
  \caption{a) An example of the sketch-based spatial search system, which allows users to draw a sketch and rank the spatial search results.. b) Some examples of sketches which are drawn using the system.}
  \label{fig:system}
\end{figure}

\subsection{Data Annotation}
To build the relationship between sketch map and spatial scene, we develop a sketch map search system based on the proposed method, which aims to collect a variety of sketch maps and allows users to rank the results regarding their technological background and expertise. Using this system (as shown in Fig. \ref{fig:system}), users can draw a digital sketch map by dragging icons onto the sketch panel, with the icons representing different types of geographic objects. The system will also record the users' intersection with the search results and annotate the results accordingly for each sketch map.

Moreover, the system described here is designed to be maintained online, allowing crowd-sourced sketches to be added continuously. As the increase of training samples over time, the quality and quantity of the training dataset are expected to improve the model performance.

\subsection{Data Processing}
\begin{figure}[h]
  \centering
  \includegraphics[width=\linewidth]{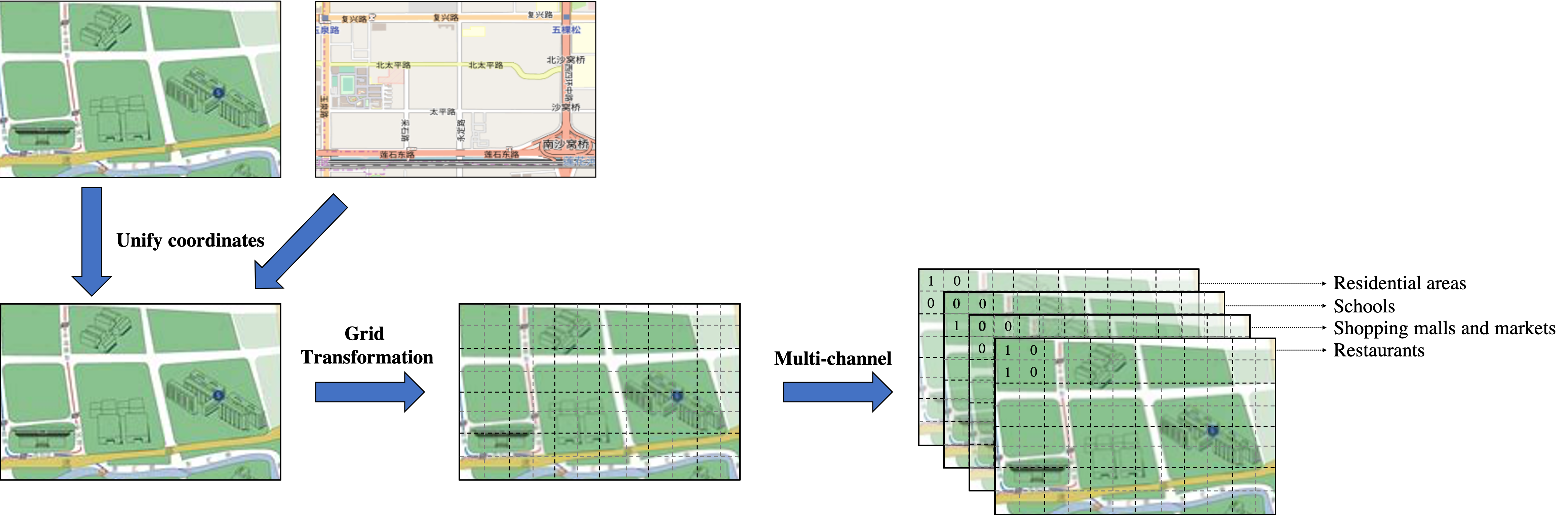}
  \caption{The flowchart of data pre-processing for spatial scenes.}
  \label{fig:data}
\end{figure}

% Unified coordinates.
The data pre-processing of DeepSSN (as shown in Fig. \ref{fig:data}) consists of three steps: coordinates unification, grid transformation and multi-channel fusion and transformation. The coordinates unification step unifies two geographic data sources into same coordination system and merges their attribute information. In the gird transformation process, the whole study area was divided into 40 pixel \begin{math}\times\end{math} 40 pixel grids with each grid representing a geographic area of 10 \begin{math}\times\end{math} 10 square meters. The map dataset in the study area is divided into 6565 spatial scenes.

% Build multi-channel data.
Furthermore, the multi-channel convolutional networks are pervasive in domains where the inputs can naturally be categorized into different channels, e.g. color channels in computer vision \citep{cheng2016person}, wave lengths in speech recognition \citep{hoshen2015speech}. In DeepSSN, we fuse the above mentioned two datasets and define 15 channels to represent spatial scenes, which are comprised by various types of representative geographic objects listed in Table \ref{tab:geographicfeatures}. We calculate the number of each type of POIs in each grid cell to transform the multi-types of POIs into multi-dimensional greyscale rasters. For example, when there is a hospital in a grid cell, the cell is set to 1 in the hospital channel, otherwise set to 0. In some cases, one cell may contain more than one POI in a specific type of POI channels (e.g., the number is $n$), and the value in that cell for that channel would be $n$. Finally, a spatial scene is transformed and fused into a multi-dimensional matrix, which is further fed into a deep learning model. In the sketch query stage, we also use the same processing steps to formulate a sketch map as a multi-dimensional matrix.

\begin{table}[!htbp]
\centering
\caption{The list of geographic map layers used in this research.}
\label{tab:geographicfeatures}
\begin{tabular}{cc}
\toprule
No.& Layer name\\
\midrule
0& Buildings (From OSM) \\
1& Land uses (From OSM) \\
2& Schools and education institutions\\
3& Hotel accommodation \\
4& Governmental agencies and institutes \\
5& Roads and stations \\
6& Greenbelt and plants \\
7& Restaurants \\
8& Residential areas \\
9& Rivers and lakes \\
10& Shopping malls and markets \\
11& Office buildings and commercial districts\\
12& Hospitals and health care providers \\
13& Life service business \\
14& Scenic spots and resorts \\
\bottomrule
\end{tabular}
\end{table}

\subsection{Data Augmentation}
Different from metric maps, the information in sketch maps may be distorted, schematized, exaggerated, and even incomplete \citep{wang2015invariant}. Such peculiarities in sketch maps is important in assessing cognitive maps and in learning invariant information for querying spatial databases \citep{schwering2014sketchmapia}. Data augmentation is a method for high-quality expansion for the quantity and diversity of training data sets. Thus, we augment the training data by simulating the variant information, which is more likely to occur in sketch maps, via these above-mentioned peculiarities of sketch maps. Specifically, we use a set of operations (e.g., dropout, shift, rotate and scale) randomly to simulate and augment sketch maps. We enlarge the size of training dataset to 20 times by using the data augment techniques. The detailed parameter settings for the data augmentation of different types of geographic objects are shown in Table \ref{tab:dataAugmentParameter}. The data enhancement process resulted in 131,300 training samples. These augmented data can be used to boost the train process and evaluate the similarity task with ground truth.

\begin{table}[!htbp]
\small
\centering
\caption{The parameter settings of data augmentation.}
\label{tab:dataAugmentParameter}
\begin{tabular}{
p{0.15\columnwidth}
p{0.08\columnwidth}
p{0.05\columnwidth}
p{0.12\columnwidth}
p{0.12\columnwidth}
p{0.15\columnwidth}
p{0.23\columnwidth}
}
\hline
{Geographic object type}& {Selection probability}& \multicolumn{5}{c}{Simulation probabilities and transformations} \\
\cline{3-7} ~& ~& Drop      & Shift      & Rotation & Scaling & Shift and Scaling \\
\midrule

Schools and education institutions& 0.3& 0.1& 0.3/$\pm 100m$ & 0.2/$\pm 45^\circ$ & 0.2/[0.2, 1.2]& 0.2/$\pm 100m$/[0.2, 1.2]\\

Hotel accommodation& 0.2& 0.2& 0.2/$\pm 100m$ & 0.2/$\pm 45^\circ$& 0.2/[0.2, 1.2]& 0.2/$\pm 100m$/[0.2, 1.2]\\

Governmental agencies and institutes& 0.2& 0.2& 0.2/$\pm 100m$ & 0.2/$\pm 45^\circ$ & 0.2/[0.2, 1.2]& 0.2/$\pm 100m$/[0.2, 1.2]\\

Roads and stations& 0.4& 0.3& 0.2/
$\pm 50m$ & 0.2/$\pm 45^\circ$ & 0.1/[0.2, 1.2]& 0.2/$\pm 50m$/[0.2, 1.2]\\

Greenbelt and plants& 0.3& 0.3& 0.2/$\pm 100m$ & 0.1/$\pm 45^\circ$ & 0.2/[0.4, 1.2]& 0.2/$\pm 100m$/[0.4, 1.2]\\

Restaurants& 0.5& 0.4& 0.2/$\pm 100m$ & 0.1/$\pm 45^\circ$ & 0.1/[0.2, 1.2]& 0.2/$\pm 100m$/[0.2, 1.2]\\

Residential areas& 0.3& 0.4& 0.2/$\pm 100m$ & 0.1/$\pm 45^\circ$& 0.1/[0.2, 1.2]& 0.2/$\pm 100m$/[0.2, 1.2]\\

Rivers and lakes& 0.3& 0.2& 0.2/$\pm 100m$& 0.2/$\pm 45^\circ$& 0.2/[0.5, 1.2]& 0.2/$\pm 100m$/[0.5, 1.2]\\

Shopping malls and markets& 0.4& 0.2& 0.3/$\pm 100m$ & 0.2/$\pm 45^\circ$ & 0.2/[0.2, 1.2]& 0.2/$\pm 100m$/[0.2, 1.2]\\

Office buildings and commercial districts& 0.4& 0.4& 0.2/$\pm 100m$ & 0.1/$\pm 45^\circ$ & 0.1/[0.2, 1.2]& 0.2/$\pm 100m$/[0.2, 1.2]\\

Hospitals and health care providers& 0.2& 0.2& 0.2/$\pm 100m$& 0.2/$\pm 45^\circ$ & 0.2/[0.2, 1.2]& 0.2/$\pm 100m$/[0.2, 1.2]
 \\

Life service business& 0.2& 0.3& 0.2/$\pm 100m$ & 0.1/$\pm 45^\circ$ & 0.2/
[0.2, 1.2]& 0.2/$\pm 100m$/[0.2, 1.2]\\

Scenic spots and resorts& 0.2& 0.2& 0.2/$\pm100m$ & 0.2/$\pm 45^\circ$ & 0.2/[0.5, 1.2]& 0.2/$\pm 100m$/[0.5, 1.2]\\

\bottomrule
\end{tabular}
\end{table}

\section{Experiment and Evaluation}

We design two experimental scenarios to evaluate the precision and scalability of the DeepSSN. The first one is to test the ability of assessing the similarity of spatial scenes generated directly by the map dataset, while the other is to test the ability of similarity assessment between user-input sketch maps and spatial scenes from the geographic database. 

In the first scenario, we built spatial scene sets from the simulated scenes with random transformation coefficients that control the shift, rotation, scaling, and other geometric operations of objects on maps, whose exclusive label corresponds to the original spatial scene in the conflated map dataset. We then randomly selected 500 simulated spatial scenes with various categories for the comparison experiments to other baseline methods.

In the second scenario, the testing spatial scenes were collected from the aforementioned sketch-based map search system, from which the scenes are generated and annotated by crowd-sourced users. To evaluate the degree of user approval for the spatial scene search results, we randomly selected 150 sketches inputted by crowd-sourced users and recorded their rankings of the results from the scene database. Additionally, we evaluate the performance of the triplet mining strategy and latent representation dimensions on the spatial scene search tasks. 

All the experiments were carried out on the AI computing cluster of the Chinese Academy of Sciences, which deploys 48 Dawning W780-G20 servers and each server with 2 cores of Intel Xeon 2650v4, NVIDIA Tesla P100 GPU, and CUDA-9.0 and cuDNN-7.30 for GPU-accelerated computation. 

\subsection{Evaluation Metrics}
Compared to MAP (Mean Average Precision), MRR (Mean Reciprocal Rank) and Precision@k (top-k precision rate) perform better in evaluation of the performance of GIR \citep{yan2017itdl}. In DeepSSN, we compare the rankings of each sketched-based spatial scene based on the augmented spatial scenes from the system and the ground truth using the Mean Reciprocal Rank (MRR) and the top-k precision rate (Precision@k) to test the performance of our method against other baseline approaches.

\textbf{Mean Reciprocal Rank}. Given the uniqueness of the sketch-based search in spatial scenes, i.e., an identity of spatial scene only corresponds to one sample among all spatial scenes, the first correct match of the relevant spatial scenes in the ranked candidate list is selected. We employ the Mean Reciprocal Rank (MRR), which is the average of first correctly matched reciprocal ranks of the results over the sample query set \citep{voorhees1999trec} and has been widely used in GIR \citep{mckenzie2015also,yan2017itdl}. The larger the MRR value, the better the performance.

\textbf{Precision@k}. For web-based geographic information retrieval, the recall rate may not be a meaningful metric, as many queries may have thousands of relevant geo-indicative textual documents and images. Instead, the precision at $k$ returned results would be an informative metric with promising computation time under such circumstances, e.g., P@10 or ``Precision at 10" is evaluated at a given cut-off rank of 10, considering only the precision of top-10 most relevant spatial scene results returned by the system. One limitation of the \textit{Precision@k} metric is that we cannot explicitly evaluate the actual rank order; we can only know how many results are relevant. Such a limitation could be mitigated by the aforementioned metric MRR.

\subsection{Performance Evaluation}
\subsubsection{Baseline comparison}

%There are three aspects to evaluate the performance of DeepSSN, the first one is from the number of dimensions of the spatial scene embedding. The second one is to compare our embedding with the hand-crafted feature embedding using the proposed evaluation schemes as a baseline. And the last one is to compare the spatial scene embedding trained from other machine learning methods and deep learning methods.

We first compare our results to a set of baseline approaches based on the map scene dataset with ground-truth labels. Those approaches are categorized into three major types: 
(1) a traditional method in spatial similarity search, i.e., the ratio between the number of objects that have been matched and the total number of objects in both scenes. 
(2) Machine learning methods, which compute the k-nearest neighbors (k-NN) for the target sample using a distance metric. Here, we use the ball tree \citep{liu2003efficient}, kd-tree \citep{shibata2003kd} and a fast brute-force search algorithm using GPU \citep{garcia2008fast} to boost the computation process. 
(3) Deep learning methods, which combine state-of-the-art deep learning models and the triplet loss to learn the distance mapping function. The retrieval of a spatial scene is similar to image retrieval approaches, most of which combine image classification methods and the triplet loss. Therefore, we choose the multilayer perceptron (MLP) \citep{ruck1990feature}, AlexNet \citep{krizhevsky2012imagenet}, DenseNet \citep{huang2017densely} and ResNet~\citep{he2016deep} with different number of layers (including 18, 34, 50 layers, respectively). The parameters of these comparative methods are all set based on the authors' suggestions in their original publications. All the compared methods based on deep learning were trained using the same triplet-mining strategy and the triplet loss framework.

\begin{figure*}[!h]
  \centering
  \includegraphics[width=\linewidth]{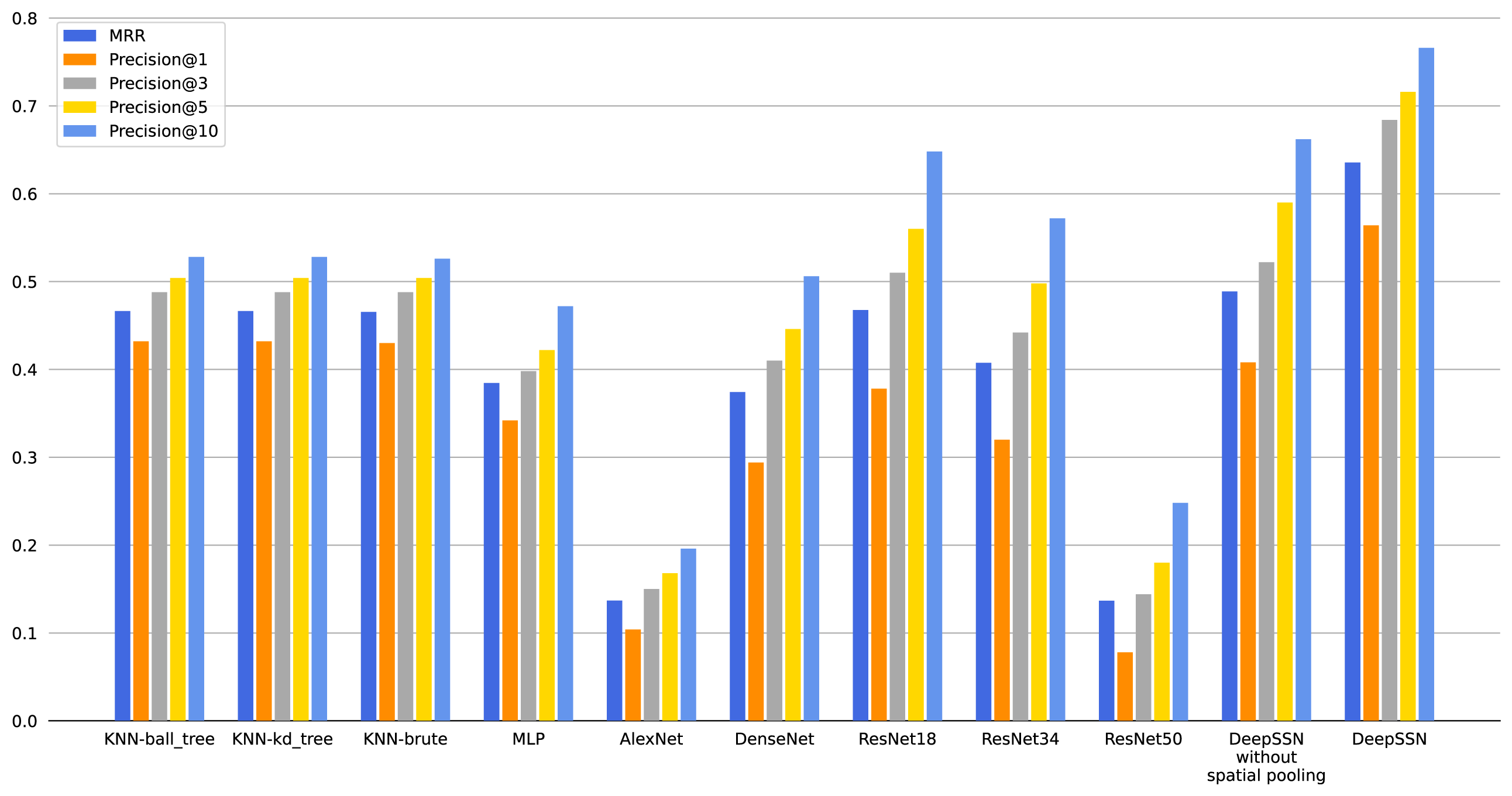}
  \subfloat(a) \\
  \includegraphics[width=\linewidth]{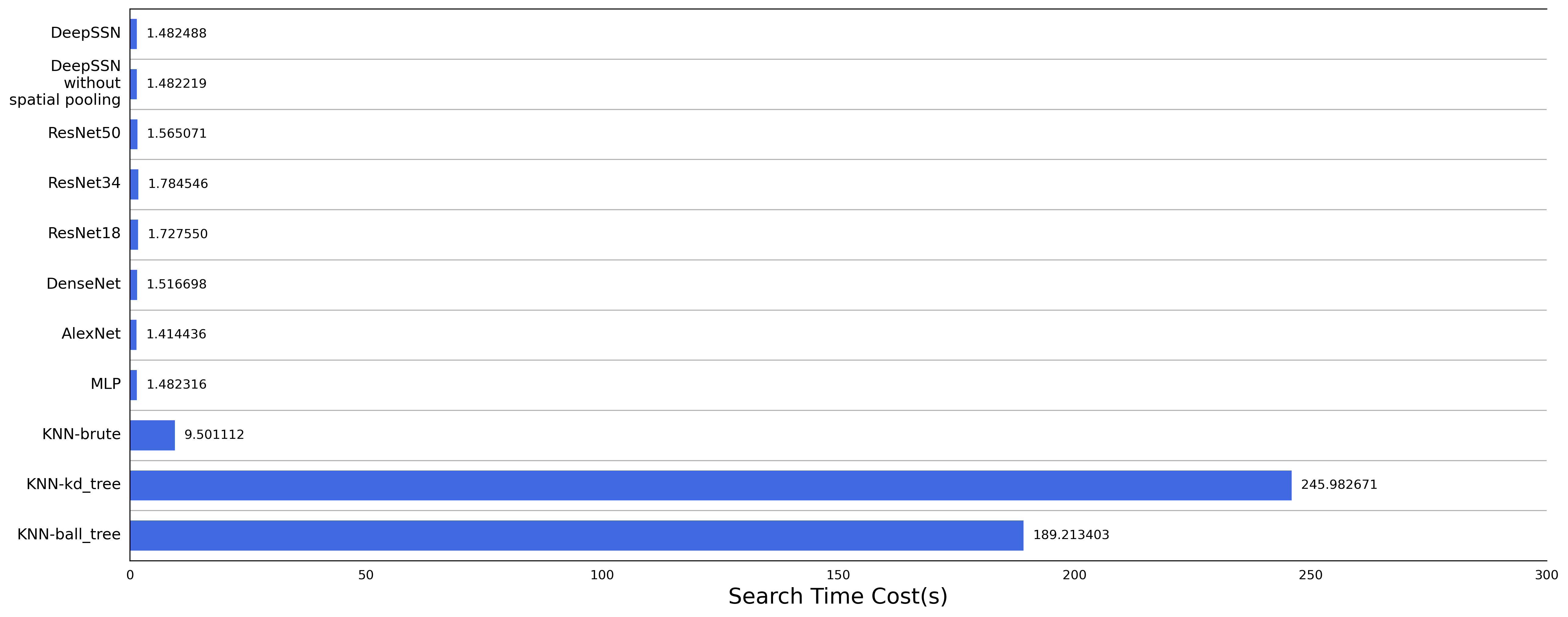}
  \subfloat(b) \\
  \caption{Comparison of of our method and the other methods. (a) Comparison of the Mean Reciprocal Rank (MRR) result and the top-k precision for the evaluation of searches on random-sample simulated sketches. Each bar is color coded for a specific evaluation metric (i.e., MRR, Precision@k). (b) Comparison of the computation time cost on searching for random-sample simulated sketches.}
  \label{fig:result}
\end{figure*}

The computation time and quality comparisons of DeepSSN with baseline approaches based on geographic object information are shown in Fig. \ref{fig:result}. First, we find that the learning of latent embeddings greatly decreases the search time compared to k-NN methods, which directly compare the distance of the sketch and spatial scenes in hand-crafted feature space. The time complexity of the traditional spatial retrieval method is \begin{math}O(T \times K \times N^M)\end{math}, where $T$ refers to the count of sketch queries, $K$ refers to the count of spatial scene candidates, $N$ is the count of geographic objects in each spatial scene, and M is the count of adjacent geographic objects.
When the value of $M$ is 5, the time cost of the traditional method is 780s and the MRR is 0.436 given the information loss of geographic objects, while DeepSSN gets a MRR of 0.641 with a computation time of less than 2s. In terms of k-NN methods, the time complexity is \begin{math}O(T \times D)\end{math}, where $T$ is count of sketch queries, $D$ is the number of dimensions, which mainly depends on $D$. Obviously, the learning of latent embeddings of spatial scene helps improve the overall performance significantly.

DeepSSN also outperforms the traditional CNN-based methods in spatial scene search task with higher MRR and precision@k values as shown in Fig. \ref{fig:result}. It demonstrates that larger and multi-scale convolutional filters in DeepSSN help capture multi-scale spatial features and increase the overall performance. All of these results show that the DeepSSN is indeed a valuable method for spatial scene similarity search. In online inference application scenarios, one would choose a suitable method regarding both model accuracy and computation time; the proposed DeepSSN method would be a good candidate.

\subsubsection{Performance comparison of different loss functions, triplet mining strategies and convolution kernel sizes}
Moreover, regarding the relation between spatial scene query and graph embedding learning, we incorporate the triplet-loss function and an triplet mining strategy in spatial similarity assessment. We conduct additional empirical studies to evaluate the impact of our strategies, where the neural network model is implemented without the triplet-loss or without the triplet hard-sample mining strategy as a comparison.

% loss 
We compare the performance of the network architecture without the triplet loss with the performance of DeepSSN. The model uses a common loss function, namely the cross-entropy loss function \citep{goodfellow2016deep}, as the baseline. The comparison results of MRR and precision@k are displayed in Table \ref{tab:loss}. We find that the triplet-loss approach outperforms the cross-entropy loss approach in spatial scene search tasks, which may be a result of the fact that the sketch map is quite sparse and the number of spatial scene categories is large. The triplet loss fits better in such situations. 

\begin{table}[!htbp]
\centering
\caption{Comparison between the cross-entropy loss and the triplet loss.}
\label{tab:loss}
\begin{tabular}{p{2.5cm}ccccc}
\hline
	& MRR&	Precision@1&	Precision@3&	Precision@5&	Precision@10 \\
\hline
Cross-entropy loss& 	0.054& 	0.050& 	0.0567& 	0.060& 	0.063 \\
Triplet loss& 	0.641& 	0.576& 	0.678& 	0.714& 	0.752 \\
\hline
\end{tabular}
\end{table}

% mining strategy 
Furthermore, we compare the performance of the network architecture without the triplet hard-sample mining strategy with the performance of DeepSSN. The random sampling strategy is used as the baseline. The comparison results of MRR and precision@k for the two strategies are shown in Table \ref{tab:strategy}. It shows that the triplet mining strategy outperforms the random sampling strategy in spatial scene search tasks as well.

\begin{table}[!htbp]
\centering
\caption{The performance comparison of different mining strategies}
\label{tab:strategy}
\begin{tabular}{p{2cm}ccccc}
\hline
	& MRR&	Precision@1&	Precision@3&	Precision@5&	Precision@10 \\
\hline
Random sampling & 	0.529&	0.452&	0.576&	0.608&	0.676 \\
Triplet mining & 	0.641& 	0.576& 	0.678& 	0.714& 	0.752 \\
\hline
\end{tabular}
\end{table}

The comparisons of the triplet-loss function and the triplet mining strategy show that this research provides an effective solution to integrate the triplet-loss function into CNNs. Furthermore, the empirical results indicate that the triplet-loss function could lead to improvement of search methods in spatial analysis by taking the spatial relations among geographical objects into consideration.

% kernel sizes and spatial pyramid pooling 
In addition, we compared the performance of DeepSSN with different convolution kernel sizes. The kernel size of [11,7,5] is used as the baseline. The comparison results of MRR and precision@k for these models are shown in Table \ref{tab:kernel}. It shows that the kernel size of [11,7,5] achieves higher accuracy and outperforms the models with other kernel sizes. Also, we compare the effect of spatial pyramid pooling with the use of a single max-pooling layer, with an illustration of the comparison shown in Fig. \ref{fig:result}. In most cases, spatial pyramid pooling (SPP) is superior to the max-pooling approach with about a \textbf{14\%} improvement in MRR, which shows that spatial pyramid pooling better captures multi-scale features in spatial scenes compared to max-pooling.

\begin{table}[!htbp]
\centering
\caption{The performance comparison of different convolution kernel sizes}
\label{tab:kernel}
\begin{tabular}{p{2cm}ccccc}
\hline
	& MRR&	Precision@1&	Precision@3&	Precision@5&	Precision@10 \\
\hline
DeepSSN \\without SPP & 	0.489&	0.408&	0.522&	0.590&	0.662 \\
DeepSSN \\kernel[7,5,3] & 	0.379& 	0.290& 	0.424& 	0.472& 	0.536 \\
DeepSSN \\kernel[5,3,1] & 	0.363& 	0.276& 	0.406& 	0.448& 	0.528 \\
DeepSSN \\kernel[13,9,7] & 	0.532& 	0.486& 	0.534& 	0.587& 	0.649 \\
DeepSSN \\kernel[11,7,5] & 	\textbf{0.641}& \textbf{0.576}& \textbf{0.678}&	\textbf{0.714}& \textbf{0.752} \\
\hline
\end{tabular}
\end{table}

Finally, we use the same method to expand the test set data size from 500 to 1000, the comparison results of MRR and precision@k for the experiments are shown in Table \ref{tab:datasize}. It shows that the amount of data does not have much impact on the final test results. It is worth noting that labeled spatial scene data are not openly available in public image data repository such as ImageNet. We would like to test more different datasets if such data are available in future.

\begin{table}[!htbp]
\centering
\caption{The performance comparison of different testing data sizes}
\label{tab:datasize}
\begin{tabular}{p{2cm}ccccc}
\hline
	& MRR&	Precision@1&	Precision@3&	Precision@5&	Precision@10 \\
\hline
Test data \\(500) & 	0.641& 	0.576& 	0.678& 	0.714& 	0.752 \\
Test data \\(1000) & 	0.635&	0.582&	0.658&	0.698&	0.750 \\
\hline
\end{tabular}
\end{table}

\subsection{User rank-based evaluation}
The above mentioned experiments are all based on computer generated spatial scenes. In order to further evaluate the performance of the proposed method based on human spatial cognition, we sampled 150 sketches in different categories of scenes as the test data. The participants involved in the search result ranking of images consist of twenty volunteers who participated the evaluation. They are members of our research group and colleagues in our university. Random sampling was adopted on the collected sketches to generate unbiased results. To evaluate the user ranking result on multi-perspective views, we randomly divided the users into two groups. 
Participants in one of the groups selected only similar spatial scenes returned from the search and ranked the scenes by the degree of similarity in the first web page (in total 12 records of searched spatial scenes), an example of which is shown in Fig. \ref{fig:system}. 
The other group of participants ranked all the scene results in the first web page by the degree of similarity between the sketch and the system searched results. Through this process, we could capture more information about the learning ability of our proposed DeepSSN by comparing the ranking result between the model and that of human judgment. 

To evaluate the precision rate of the method, the empirical study of the first group is implemented based on 100 queries from over ten users. Table \ref{tab:userTable} shows the result of the user rank-based evaluation, which derives the hitting proportion of the user rank order with the model learned order. The value reported in the row $i$ and the column $j$ is computed as $a_{ij}=\frac{1}{n}n_{ij}$, where $n_{ij}$ is the number of queries whose order ranked by the model is $i-th$ and considered by a user to be in the $j-th$ similar result. And $n$ is the number of all queries. The sum of the column values should be less than 1. Table \ref{tab:userTable} shows that the search and retrieval results of DeepSSN are highly related to the sketches drawn by users. For instance, 0.13 in the row 1 and the column 1 means that 13\% of the results are most relevant to the sketch queries and both the model and the user rank them as the top-1 candidate. 0.18 in the row 1 and the column "Total" means that 18\% of the results ranked as the most relevant to the sketch queries by the model are also shown in the users' top-3 ranking results. There are some noticeable observations where our method had a precision of 0.77 from the twelve spatial scenes recommended by our system, and indicating that the returned spatial scenes by the DeepSSN model are very similar to the sketch queries, according to the user judgment.

Additionally, the normalized discounted cumulative gain (nDCG) measure is also used in our empirical study, which evaluates the average performance of a ranking algorithm by summing the true scores ranked in the order and applying a logarithmic discount factor on result relevance values \citep{jarvelin2002cumulated}. The value of nDCG for these queries is 0.626, which indicates the system search results are strongly related to user sketch queries \citep{mcsherry2008computing}. In short, DeepSSN shows the ability to achieve high quality latent representation of spatial scenes and return multiple similar spatial scenes.

In order to investigate the ranking diversity among different users, the empirical study of the second group was implemented based on 50 queries from the users. We use the Kendall's tau, which is a measure of the concordance between different rankings \citep{kendall1945treatment,gao2017data}, to assess the agreement among the rankings for the searched spatial scene results of those users. If the two sequences are completely accordant, the value of Kendall's tau is 1. On the contrary, values close to -1 indicate strong disagreement between different rankings. The average Kendall's tau of the 50 queries from the users was 0.639. Besides, the two-sided p-value for the null hypothesis test was 0.033. The Kendall's tau and the p-value indicate that there is a significant positive agreement among the users' ranking sequences at a significance level of 0.05.

% Change the intercolumn space
% \setlength{\tabcolsep}{4pt}

\begin{table}[!htbp]
\centering
\caption{A comparison of the model ranking result and the user evaluation result.}
\label{tab:userTable}
\begin{tabular}{|c|c|c|c|c|}
\hline
{Model Rank} & \multicolumn{3}{c|}{User Rank} & Total \\ \cline{2-5} 
                            & 1st      & 2nd      & 3rd      &       \\ \hline
\begin{math}1^{st}\end{math}&	0.13&	0.04&	0.01&	0.18 \\
\begin{math}2^{nd}\end{math}&	0.06&	0.05&	0.01&	0.12 \\
\begin{math}3^{rd}\end{math}&	0.06&	0.05&	0.01&	0.12 \\
\begin{math}4^{th}\end{math}&	0.04&	0.05&	0&	0.09 \\
\begin{math}5^{th}\end{math}&	0.03&	0.04&	0.03&	0.10 \\
\begin{math}6^{th}\end{math}&	0.08&	0.01&	0.03&	0.12 \\
\begin{math}7^{th}\end{math}&	0.08&	0.03&	0.01&	0.12 \\
\begin{math}8^{th}\end{math}&	0.06&	0.03&	0.02&	0.11 \\
\begin{math}9^{th}\end{math}&	0.06&	0.07&	0&	0.13 \\
\begin{math}10^{th}\end{math}&	0.07&	0.04&	0.01&	0.12 \\
\begin{math}11^{th}\end{math}&	0.06&	0.04&	0.03&	0.13 \\
\begin{math}12^{th}\end{math}&	0.04&	0.06&	0.03&	0.13 \\
Total&	0.77&	0.51&	0.19& ~ \\
\hline
\end{tabular}
\end{table}

\begin{figure}[!htb]
  \centering
  \includegraphics[width=\linewidth]{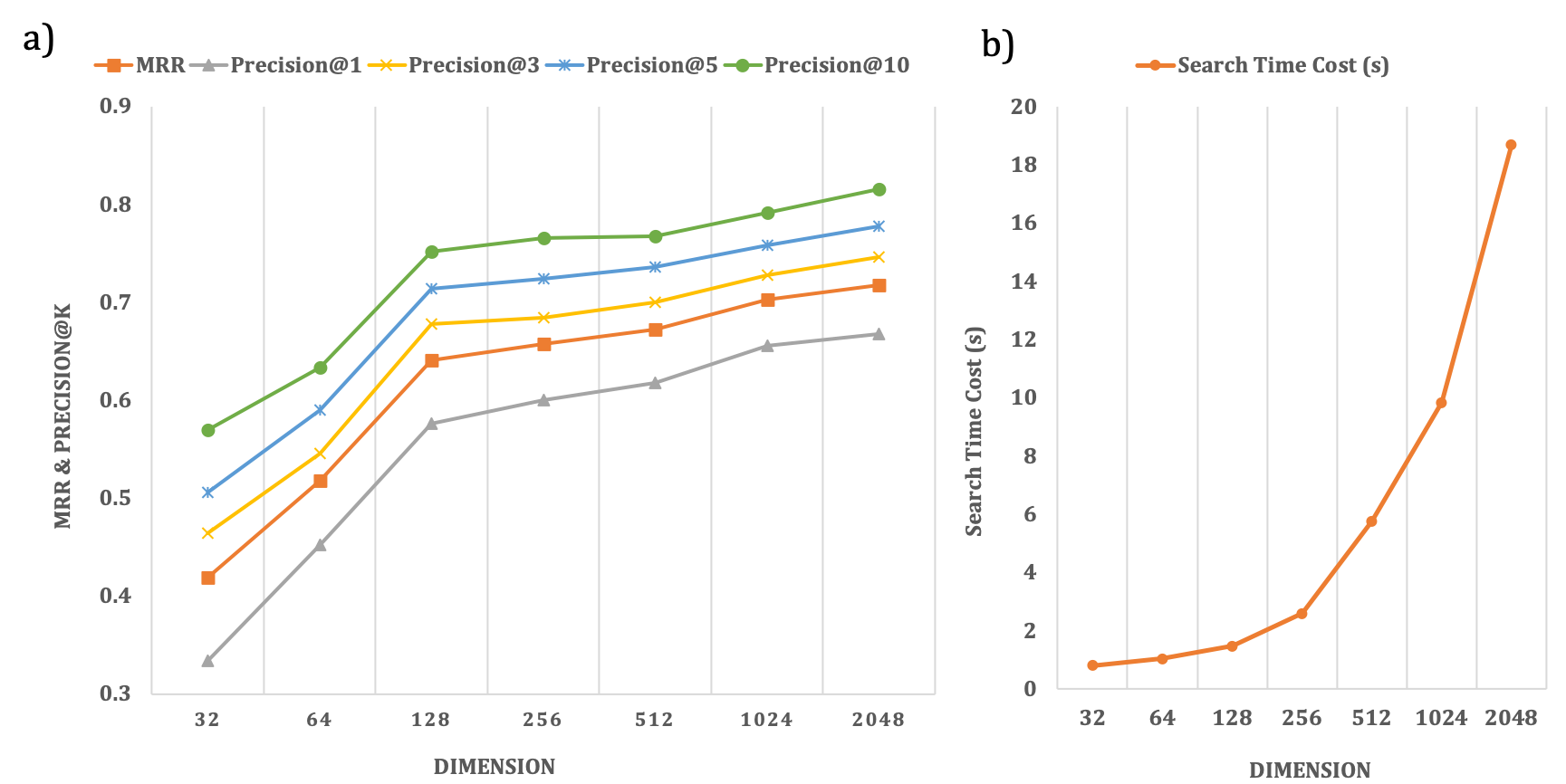}
  \caption{(a) Mean Reciprocal Rank(MRR) and precision@k for dimension evaluation. (b) Search time cost comparison at different embedding dimensions.}
  \label{fig:dimension}
\end{figure}

\section{Discussion} \label{sec:disscussion}

\subsection{Selection of embedding dimensions}  \label{sec:dimensions}
In application scenarios, spatial scene similarity based search, such as in way-finding and criminal investigation, needs to accomplish the task of spatial scene matches between a sketch map and a candidate scene quickly. The number of dimensions for the embedding vectors should be limited since large dimensional data generally need more computation time. We select 128 dimensions for all the experiments, which is ideal for large-scale data computation, as proved in the FaceNet \citep{schroff2015facenet}.  Meanwhile, by changing the embedding dimensionality to various scales, we compare the performance of the model and report the result in Fig. \ref{fig:dimension} to check model sensitivity. The experiment shows that after 128-dimensions the precision grows slowly, while the computation time increases rapidly. As a system being deployed online, the time cost of search is more critical than that in offline systems. The search time as an efficiency measure influences the users’ experience. One would expect that a larger dimensional embedding could perform better than a smaller one with higher precision. However, a larger dimensional embedding requires more training and more processing time. Therefore, as a tradeoff, a smaller dimension of embedding might be utilized with a minor loss of precision but could be deployed on the Web and mobile devices due to their memory and computation limits.

\subsection{Sparsity of sketches}
\begin{figure}[!htb]
  \centering
  \includegraphics[width=\linewidth]{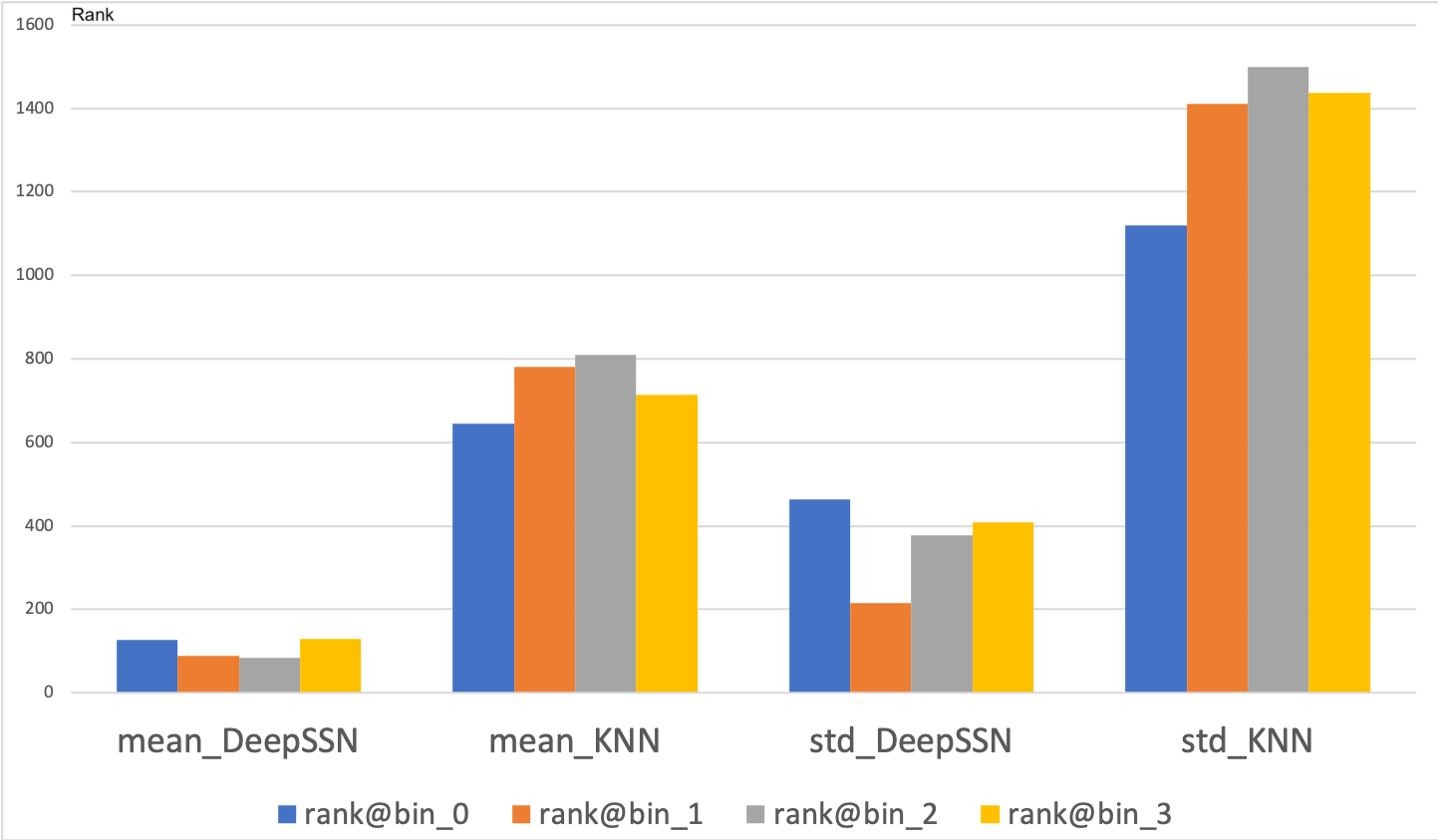}
  \caption{The comparison of the rank order of the DeepSSN model against the KNN model on the sparsity of sketches.}
  \label{fig:sparse}
\end{figure}

To evaluate the impact of sketch sparsity (i.e., zeros in a matrix), we compute the sparsity level of the sampled sketch scenes based on their embeddings (i.e., multidimentional vectors derived from Section 3.1)~\citep{aharon2006k}. According to the sparsity quartile, we divide the sampled sketches into four bins ordered from sparse to the dense, namely, ``bin\_0'', ``bin\_1'', ``bin\_2'', and ``bin\_3''. The comparisons between DeepSNN and KNN methods based on the sparsity of sketch queries are shown in Fig. \ref{fig:sparse}. Lower mean rank orders and lower standard deviation of ranks indicate better model performance of the returned scene results. The result shows that the rank order for the retrieved results of DeepSSN outperforms KNN in both the mean value of rank order and its standard deviation. In DeepSSN, the rank orders of the most sparse (``bin\_0'') and the most dense bins (``bin\_3'') are higher than that in ``bin\_1'' and ``bin\_2'', which demonstrate that the sparsity and the density of sketches indeed have certain degree of influence on the retrieval results.

\section{Conclusion and future work}
In this study, we propose a novel DeepSSN model to assess spatial scene similarity and support sketch-based geographic information retrieval. The triplet loss function and an efficient hard-triplet mining strategy are employed to facilitate the deep learning training process from large amounts of training data and boost the convergence rate. The empirical results demonstrate that the DeepSSN model outperforms traditional CNN-based methods and other baseline approaches in both precision and convergence time. The sketch-based spatial query system based on the DeepSSN model performs effectively and precisely than baselines. This research offers insights into how image-based GIR research will progress with the advancement of deep learning in the emerging field of GeoAI. Specifically, it suggests that a combination of spatial pyramid pooling, the triplet-loss function, and the hard-triplet mining strategy using qualitative constraint networks may be a promising tool for deep learning approaches involving spatial relations.

There are three possible directions for the improvements in DeepSSN. First, Using eye-tracking equipment to track the users' interaction with candidate spatial scenes and reduce the gap between human spatial recognition process and feature-engineering approach in machine learning. Second, the extension of our experiment to more cities with various urban spatial structures could avoid over-fitting issue which is a common deficiency of CNN. Last but not least, we would like to try a combination of image features and multi-source spatial contexts (e.g., co-location patterns of geographical objects) for improving performance in spatial query tasks, which is a promising research direction in GIR and GeoAI for geographic knowledge discovery \citep{yan2018xnet,janowicz2020geoai}.

\section*{Data and codes availability}
The data and codes that support the findings of this study are available in figshare.com  with the identifier 
\url{https://doi.org/10.6084/m9.figshare.10025060.v1} 
%\URL{https://figshare.com/s/66342407c1cf2a8e8a84}.

\section*{ACKNOWLEDGMENT}
%To be added in the final proof.
The authors would like to thank Jake Kruse (University of Wisconsin-Madison) for his help on polishing the language of this paper. Danhuai Guo would like to thank the support of this research by National Key R\&D Program of China (No. 2019YFF0301300), Fundamental Research Funds for the Central Universities (BUCTRC:202132) and National Natural Science Foundation of China (No. 41971366, 91846301). 

%The authors would like to thank Jake Kruse (University of Wisconsin-Madison) for his help on polishing the language of this paper. Danhuai Guo and Yangang Wang would like to thank the support of this research by National Natural Science Foundation of China $(No.41971366,41371386,71904095)$ , National Key R\&D Program of China $(No.2018YFC0809700, 2017YFC0803300)$ and Beijing Municipal Natural Science Foundation$(No.9172023,9194027)$.

\bibliographystyle{tfv}
\bibliography{article}

\end{document}